\documentclass[10pt, a4paper]{article}

\usepackage{lrec-coling2024} 

\usepackage{natbib}
\usepackage{multibib}
\makeatletter
\def\@mb@citenamelist{cite,citep,citet,citealp,citealt,citepalias,citetalias}
\makeatother
\newcites{languageresource}{~}

\usepackage{graphicx}
\usepackage{tabularx}
\usepackage{soul}
\usepackage{inconsolata}
\usepackage{caption}
\usepackage{graphicx}
\usepackage{amsmath}
\usepackage{amsthm}
\usepackage{booktabs}
\usepackage[linesnumbered,ruled,vlined]{algorithm2e}
\usepackage{float} 
\usepackage{subfigure} 
\usepackage{enumitem}
\usepackage{amsfonts}
\usepackage{bbm}
\usepackage{multirow}
\usepackage{booktabs}
\usepackage{lipsum}
\usepackage{etoolbox}
\usepackage{graphicx}
\usepackage{color}
\usepackage{tcolorbox} 

\usepackage{xcolor}
\usepackage{hyperref}
 \definecolor{darkblue}{rgb}{0, 0, 0.5}
  \hypersetup{colorlinks=true, citecolor=darkblue, linkcolor=darkblue, urlcolor=darkblue}

\usepackage{xstring}
\usepackage{cleveref}
\usepackage{color}

\newcommand\blfootnote[1]{%
  \begingroup
  \renewcommand\thefootnote{}\footnote{#1}%
  \addtocounter{footnote}{-1}%
  \endgroup
}

\title{Unveiling Project-Specific Bias in Neural Code Models}

\name{Zhiming Li$^{1}$, Yanzhou Li$^{1}$, Tianlin Li$^{1,\dagger}$, Mengnan Du$^{2}$, Bozhi Wu$^{1}$\\ {\bf \large Yushi Cao$^{1}$, Junzhe Jiang$^{3}$, Yang Liu$^{1}$}} 

\address{$^{1}$Nanyang Technological University\quad 
         $^{2}$New Jersey Institute of Technology\\ 
         $^{3}$Hong Kong Polytechnic University\\
         \{zhiming001, yanzhou001, tianlin001, bozhi001, yushi002\}@e.ntu.edu.sg \\
         mengnan.du@njit.edu, junzhe.jiang@connect.polyu.hk, yangliu@ntu.edu.sg
         }

\abstract{
Deep learning has introduced significant improvements in many software analysis tasks. Although the Large Language Models (LLMs) based neural code models demonstrate commendable performance when trained and tested within the intra-project independent and identically distributed (IID) setting, they often struggle to generalize effectively to real-world inter-project out-of-distribution (OOD) data. In this work, we show that this phenomenon is caused by the heavy reliance on project-specific shortcuts for prediction instead of ground-truth evidence. We propose a Cond-Idf measurement to interpret this behavior, which quantifies the relatedness of a token with a label and its project-specificness. The strong correlation between model behavior and the proposed measurement indicates that without proper regularization, models tend to leverage spurious statistical cues for prediction. Equipped with these observations, we propose a novel bias mitigation mechanism that regularizes the model's learning behavior by leveraging latent logic relations among samples. Experimental results on two representative program analysis tasks indicate that 
our mitigation framework can improve both inter-project OOD generalization and adversarial robustness, while not sacrificing accuracy on intra-project IID data. Our code is available at \url{https://anonymous.4open.science/r/BPR_code_bias-BF28/README.md}.
 \\ \newline  }

\makeatletter
\DeclareRobustCommand\onedot{\futurelet\@let@token\@onedot}
\def\@onedot{\ifx\@let@token.\else.\null\fi\xspace}
\def\eg{e.g\onedot,\xspace} \def\Eg{\emph{E.g}\onedot,\xspace}
\def\ie{i.e\onedot,\xspace}

\def\etal{et al\onedot}
\makeatother
\usepackage{tikz}

\begin{document}

\maketitleabstract

\section{Introduction}
Neural network models\blfootnote{$\dagger$ Corresponding author} have revolutionized the software engineering community by achieving significant improvements on many benchmarks, while not requiring much domain expert knowledge and manual efforts. The Transformer architecture-based Large Language Models (LLMs) are nowadays the most prevalent neural code models by demonstrated to be effective on many downstream tasks. Concretely, the encoder-only LLMs have achieved improvements in many program analysis tasks (\eg~bug detection, clone detection, etc.~\cite{feng2020codebert,DBLP:conf/iclr/GuoRLFT0ZDSFTDC21}), the encoder-decoder LLMs~\cite{ahmad2021unified,raffel2020exploring,wang2021codet5}, as well as the decoder-only LLMs~\cite{chen2021evaluating,touvron2023llama}, are especially useful for sequential prediction tasks (\eg~code summarization~\cite{al2023extending,gu2022assemble}, etc.).
In particular, for the encoder-only LLMs, the CodeBert~\cite{feng2020codebert} model and its variants~\cite{DBLP:conf/iclr/GuoRLFT0ZDSFTDC21,DBLP:conf/acl/GuoLDW0022} have surpassed many delicately designed model architectures without much inductive bias thanks to the power of the pretraining and fine-tuning~\cite{devlin2018bert,brown2020language} training paradigm. 

Despite the success reported in the literature, we notice that most of these neural code models are evaluated merely under the intra-project independent identically distributed (IID) data-split setting~\cite{zhou2019devign,allamanis2020typilus}, \ie~collect code samples (often at the function level) from multiple projects, then randomly shuffle and split the dataset for training and test. However, in real-world scenarios, neural code models should be trained and tested in the inter-project setting for the majority of cases, \ie~trained on samples from a fixed set of projects while tested on samples from previously unseen projects. It is obvious that the inter-project setting is much more challenging since the vocabulary within different projects varies considerably because the naming conventions among developers differ. Thus, the real-world inter-project evaluation setting could be considered as out-of-distribution (OOD), due to the significant amount of usage of out-of-vocabulary words~\cite{bojanowski2017enriching,hu2019few} and 
different programming styles. In particular, for the representative program analysis tasks we evaluated in this paper, neural code models that achieve decent performance on the intra-project data suffer from a significant performance drop on the inter-project data. In addition, previous empirical analysis indicates that neural code models are also sensitive to semantic-preserving adversarial attacks~\cite{li2020bert} such as variable renaming and dead code insertion~\cite{yefet2020adversarial,bielik2020adversarial}. 

In this work, we aim to explore the reason why the encoder-only LLMs-based neural code models have low generalization ability on inter-project data and why they are vulnerable to naive adversarial attacks. Toward this end, we probe the model behavior with a DNN model explanation algorithm: integrated gradient~\cite{sundararajan2017axiomatic}, 
\begin{figure}[t] 
\centering 
\includegraphics[width=0.45\textwidth]{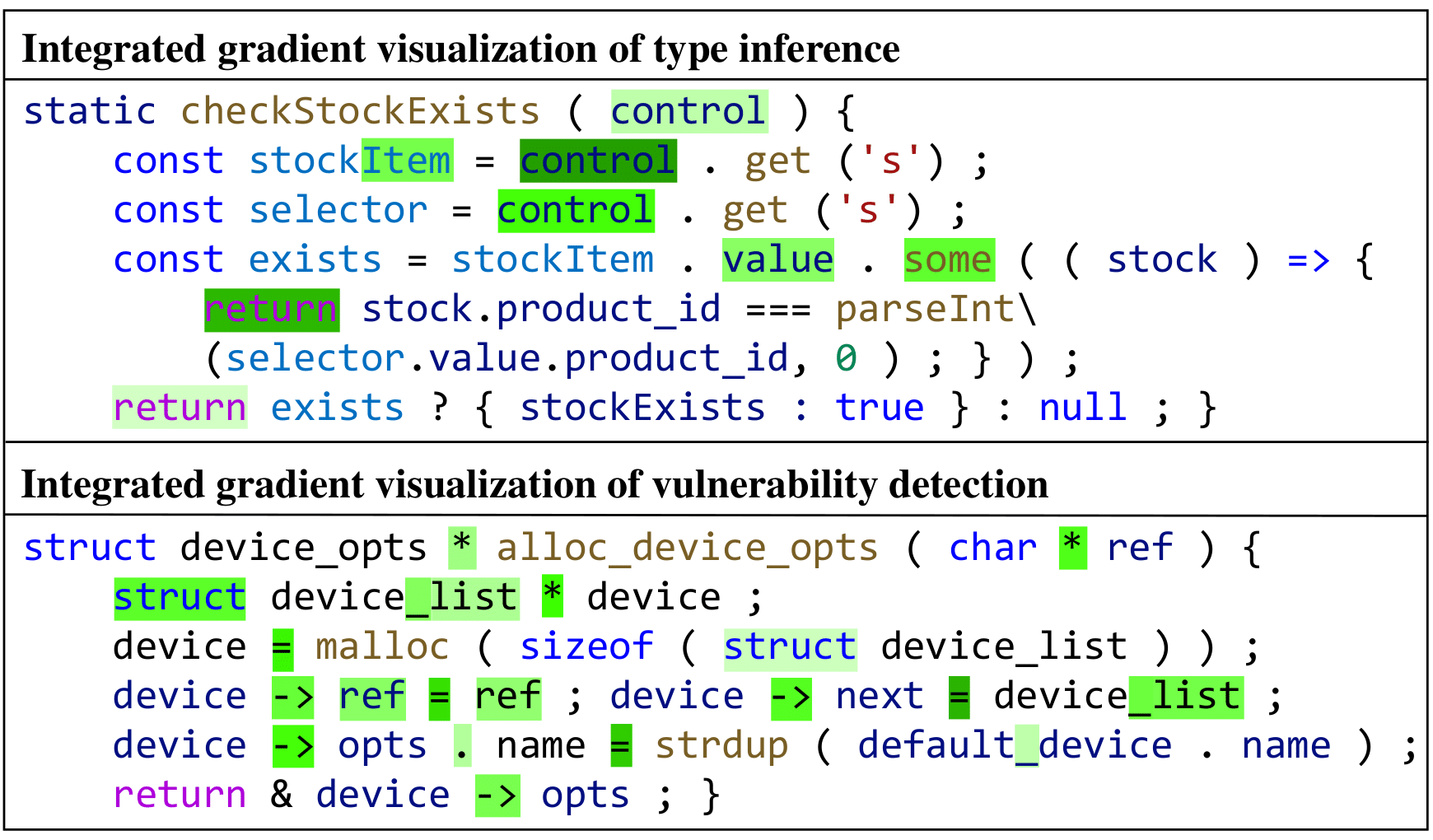} 
\caption{Illustrative examples of attribution vectors in terms of integrated gradient for cases of type inference and vulnerability detection. The shades of green indicate the weight value of the respective token in the attribution vector.} 
\label{vd_ti_case} 
\end{figure}
and find that the models heavily rely on ungeneralizable project-specific cues for prediction while ignoring ground-truth evidence when trained without regularization. We formulate it as project-specific bias learning behavior. As shown in the examples of type inference (a task that aims to infer type for variables of an optionally-typed language) and vulnerability detection (a task that aims to predict whether a code snippet contains vulnerability) in Figure~\ref{vd_ti_case} (see Section~\ref{task} for detailed descriptions of the two tasks). To predict the type of variable \texttt{stockExists}, the GraphCodeBERT (GCB) model relies primarily on uninformative sub-words such as \texttt{control}, \texttt{Item} from irrelevant variable names in the snippet. In contrast, human developers would infer based on the Boolean constant \texttt{true} in the declaration statement and infer it as a Boolean. Similarly, for the case of vulnerability detection, to predict whether the snippet contains a vulnerability, the model distributes almost all its weights to the user-defined function/variable names, while ignoring the relevant memory management API \texttt{malloc}. The vulnerability can be easily identified based on the fact that the \texttt{malloc} function is not used along with a memory deallocation operation. This learning behavior is problematic when applying the model under the inter-project setting or adversarial setting, since the semantics of these user-defined variable/function names are inconsistent across projects. 
Furthermore, we show that project-specific bias learning behavior can be interpreted with a proposed measurement termed as conditional inverse document frequency (Cond-Idf), which measures the relatedness of a token with a label and its project specificness. We observe that when trained without regularization, the model would rely heavily on the tokens that frequently co-occur with a label yet are highly semantically inconsistent and ungeneralizable for prediction.


Furthermore, for the prevalent bias mitigation methods we evaluated in this work, we observe that though these methods manage to mitigate the model from using observed shortcuts\footnote{we use bias and shortcut interchangeably in this paper.}, there is no guarantee that the post-mitigated model would infer based on the expected behavior instead of resorting to other unexpected bias. To handle this concern, we propose a novel bias mitigation mechanism, termed as BPR (\underline{B}atch \underline{P}artition \underline{R}egularization). The proposed regularization is based on the principle of invariant risk minimization (IRM)~\cite{arjovsky2019invariant}, which explicitly regularizes the model behavior by identifying common logic properties among samples. Concisely, the BPR first unshuffles and sorts samples in the training dataset according to a measure that embeds prior knowledge about logic relations. The training dataset is then divided into batches of \emph{environments} in which the samples are the most closely correlated. The in-batch representations are expected to be regularized during gradient update such that logically correlated samples would share similar representations instead of embedding other unknown shortcuts after debiasing the known ones.
The major contributions of our work are summarized as follows:
\begin{itemize}[leftmargin=*]
    \item We unveil that the previous state-of-the-art encoder-only LLMs-based neural code models trained under the intra-project setting would suffer from considerable performance drop on real-world inter-project/adversarial data, and show that this phenomenon can be attributed to the project-specific bias learning behavior.
    \item We indicate that the project-specific bias learning behavior can be interpreted with a proposed measurement called Cond-Idf.
    \item We propose a novel shortcut mitigation method, called BPR. The idea is to explicitly regularize the model behavior during training by identifying common logic properties among samples.
    \item Experimental results on two representative program analysis tasks validate that BPR can effectively improve inter-project OOD generalization and adversarial robustness while not sacrificing accuracy on intra-project IID data.
\end{itemize}
\section{Methodology}
In this section, we first introduce the analysis and interpretation methods of the project-specific bias learning behavior. Then we illustrate the details of the proposed bias mitigation mechanism, called batch partition regularization (BPR).


\subsection{Behavior Analysis and Interpretation}



We consider two typical software analysis tasks: vulnerability detection and type inference (see Section~\ref{task} for a detailed description of the two tasks). The backbone encoder-only LLMs-based models we analyze are the pre-trained CodeBERT~\cite{feng2020codebert} and GraphCodeBERT~\cite{DBLP:conf/iclr/GuoRLFT0ZDSFTDC21,DBLP:conf/acl/GuoLDW0022} models. Although they have been reported to achieve state-of-the-art performance on multiple software analysis tasks~\cite{DBLP:conf/iclr/GuoRLFT0ZDSFTDC21,DBLP:conf/acl/GuoLDW0022}, we find that compared to their high performance on the intra-project IID data, their performance drops considerably on the real-world inter-project OOD set and adversarial set for the two evaluated tasks. Intuitively, to robustly infer, the model should learn to embed abstract, generalizable code semantics, instead of using merely low-level self-defined variable/function names. In the following, we unveil that neural code models may utilize ungeneralizable tokens as shortcuts, while disregarding ground-truth evidence for predictions.
This is because these tokens co-occur frequently with the label while they are project-specific and ungeneralizable due to the developer's idiosyncrasies, which results in models' poor generalization and robustness. 

\begin{figure}[t] 
\centering 
\includegraphics[width=0.40\textwidth]{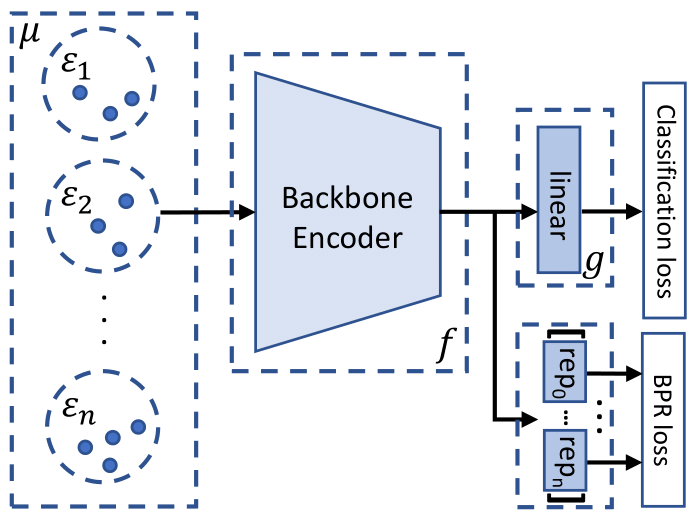} 
\caption{Overview of the proposed bias mitigation method. Training samples $x\in \mathcal{X}$ are first embedded with $\mu$ and sorted in terms of the similarity measure $\kappa$, then batchified and partitioned into multiple \emph{environments} $\varepsilon \in \mathcal{E}$ according to their labels and similarity scores. Finally, the batch partition regularization (BPR) loss is computed along with the classification loss.} 
\label{model} 
\vspace{-1em}
\end{figure}

\paragraph{Model Behavior Analysis.} We analyze the model's behavior using a post-hoc DNN model explanation algorithm: Integrated Gradient~\cite{sundararajan2017axiomatic}. Intuitively, the algorithm attributes the prediction to the input by giving each feature an importance score~\cite{montavon2018methods}, which indicates its contribution to the output. The detailed formula is as follows:

\begin{equation} \label{eq:1}
\small
    \mathrm { IGs}(x_{i})=\left(x_{i}-x_{i}'\right) \cdot \frac{1}{m}\sum_{k=1}^{m} \frac{\partial f_{y}\left(x_{i}'+\frac{k}{m}\left(x_{i}-x_{i}'\right)\right)}{\partial x_{i}}
\end{equation}
Specifically, given an input sequence with $T$ words $x_i=\{x_i^{t}\}_{t=1}^{T}$, $x_i^{t}\in \mathbb{R}^{d}$ denotes a word embedding with $d$ dimensions, the model $f(\cdot)$ outputs the prediction probability $f_{y}(x_i)$ for the ground truth label $y$, $m$ denotes the number of intermediate samples over the straightline path from baseline reference vector $x_{i}'$ to the input $x_i$. We then compute the gradients of $f_{y}(x_i)$ with respect to each input word embedding within $x_i$ and reduce each vector of the gradients to a single attribution value with the L2 norm.
We use zero word embedding as the baseline reference vector $x_{i}'$. Eventually, we obtain a feature importance vector $\mathrm {IGs}(x_{i}) \in \mathbb{R}^{T}$, where each scalar within the vector indicates the contribution of the corresponding word to $f_{y}(x_i)$.

\paragraph{Skewed Dataset Distribution.} To reflect the project-specific 
bias learning behavior, we propose a measurement called conditional inverse document frequency (Cond-Idf), which measures the co-occurrence between a word $w$ and a label $l$ (\texttt{co-occur(}$w,l$\texttt{)}), as well as its specificness across projects $\Pi$ (\texttt{specific(}$w,\Pi$\texttt{)}). The measurement is denoted as follows:
\begin{equation}
\small
     \begin{aligned}
\text { Cond-Idf }(w,l,\Pi)&=\texttt{co-occur(}w,l\texttt{)}\wedge\texttt{specific(}w,\Pi\texttt{)}\\
&=p(l \mid w) \cdot \operatorname{Idf}(w, \Pi) \\
&=p(l \mid w) \cdot \log \frac{N}{|\{\pi \in \Pi: w \in \pi\}|}
\end{aligned}
\end{equation}
\begin{algorithm}[t]
\small
\caption{BPR mitigation mechanism}\label{algorithm}
\label{alg:BPR}
\KwData{Training set $\mathcal{X}$, similarity function $\kappa$, embedding function $\mu$,  encoder $f$}
\tcp{calculate similarity matrix}
$\mathcal{X_{\mu}}\leftarrow ()$\;
\For {$(x_{i}, x_{j})\in \mathcal{X} \times \mathcal{X}$} {$\lambda_{ij}\leftarrow \kappa{(\mu(x_{i}), \mu(x_{j}))}, \lambda \in \mathbb{R}^{ |\mathcal{X}| \times |\mathcal{X}|}$}

\tcp{training set unshuffling}
\For{$\lambda_{ij} \in sorted(vec(\lambda \backslash \{\lambda_{ij}|i=j\}))$}{
\For{$k\in \{x_i, x_j\}$}{
\If{$k\notin \mathcal{X_{\mu}}$}{$\mathcal{X_{\mu}}\leftarrow \mathcal{X_{\mu}}||k; $}
}
}
\tcp{train with BPR loss}


\For{$B\in batches(\mathcal{X_{\mu}})$}
{
$\mathcal{L}_\mathrm{BPR} \leftarrow \mathbb{E}_{(x_i, x_j) \sim B\times B}{\mathbbm{1}^{ij}{\kappa(\mu(x_i), \mu(x_j))\cdot }}$\\

$\quad\quad \mathrm{S_c}(f(x_i; \theta), f(x_j; \theta))$ \\

$\mathcal{L}\leftarrow \mathcal{L}_\mathrm{DEBIAS} + \gamma_{p}\cdot\mathcal{L}_\mathrm{BPR}$

\tcp{update model weights}

$\theta \leftarrow \theta-\eta \nabla_{\Theta}\mathcal{L}$

}
\end{algorithm}
We approximate the conjunction with product t-norm~\cite{esteva2001monoidal}. Specifically, the conditional probability is calculated as $p(l|w)=\frac{count(w,l)}{count(w)}$, and $N$ is the total number of projects in the corpus: $N=|\Pi|$. We normalize the Idf term such that both the two measurements are scaled between $[0, 1]$. For each label $l$, we obtain a distribution in terms of all words in the vocabulary. Words with high Cond-Idf values indicate that they frequently co-occur with the label and are also highly project-specific. We observe that a large portion of this part of the words is the user-defined components. Although these words strongly correlate with the label in the training set, their semantics are inconsistent and ungeneralizable. For example, consider a case in type inference, a model might correlate a token \texttt{temp} with the semantics of a Boolean object in one project, since it is frequently used in cases such as \texttt{temp=True;}. However, if the model learns this spurious correlation, it might arbitrarily infer an integer variable \texttt{temp} as Boolean when dealing with \texttt{temp=1.0;} since it bases its prediction heavily on token \texttt{temp} while ignoring the ground truth evidence \texttt{1.0}.

To interpret the bias learning behavior quantitatively, we evaluate the alignment between the model's integrated gradient distribution and its corresponding Cond-Idf distribution. We first calculate the integrated gradient importance vector for every sample in the IID test set. Afterward, we calculate the mean integrated gradient value for every token in the test set vocabulary and sort them in descending order. We then use polynomial regression to approximate its corresponding Cond-Idf distribution and measure its correlation with the integrated gradient distribution.


\subsection{Proposed Mitigation Mechanism}
Although many prevalent bias mitigation baselines manage to improve generalization and robustness by removing known bias, there is no guarantee that the debiased model would infer based on the expected behavior of developers instead of resorting to other unknown biases~\cite{yoo2021towards}. Motivated by this concern, we propose a novel mitigation mechanism called batch partition regularization (BPR) to regularize the behavior of neural code models (see \Cref{model}). BPR follows the \emph{invariant-risk-minimization} (IRM)~\cite{arjovsky2019invariant} philosophy and aims to constrain neural code representation so that the model's learning behavior is expected to be invariant when handling samples with similar syntactic and semantic evidence. Details of the BPR algorithm are shown in~\Cref{alg:BPR}.
\begin{figure*}
\centering     
\subfigure{\label{fig:vd_ig_cb}\includegraphics[width=0.245\textwidth]{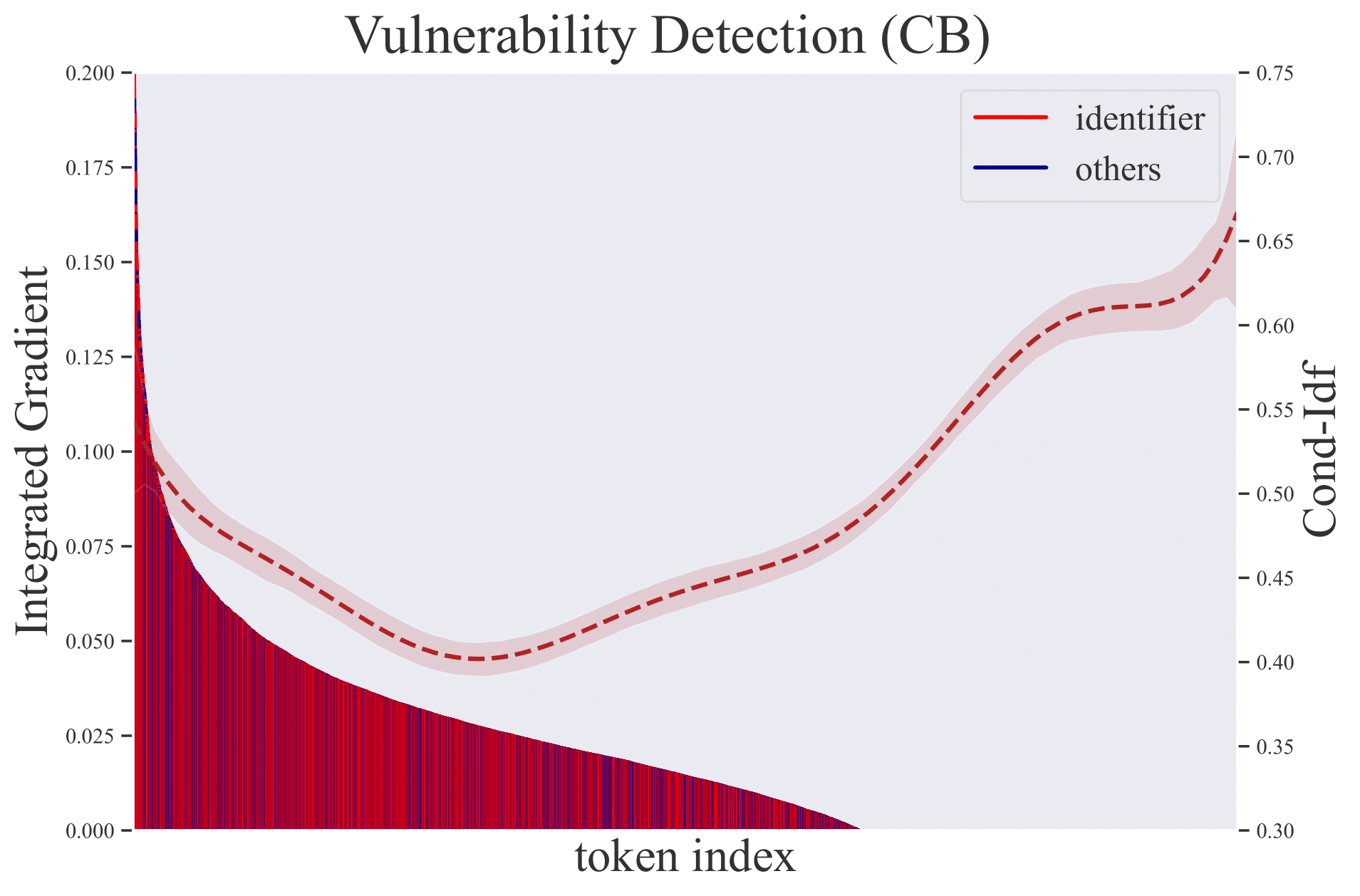}}
\subfigure{\label{fig:vd_ig_gcb}\includegraphics[width=0.245\textwidth]{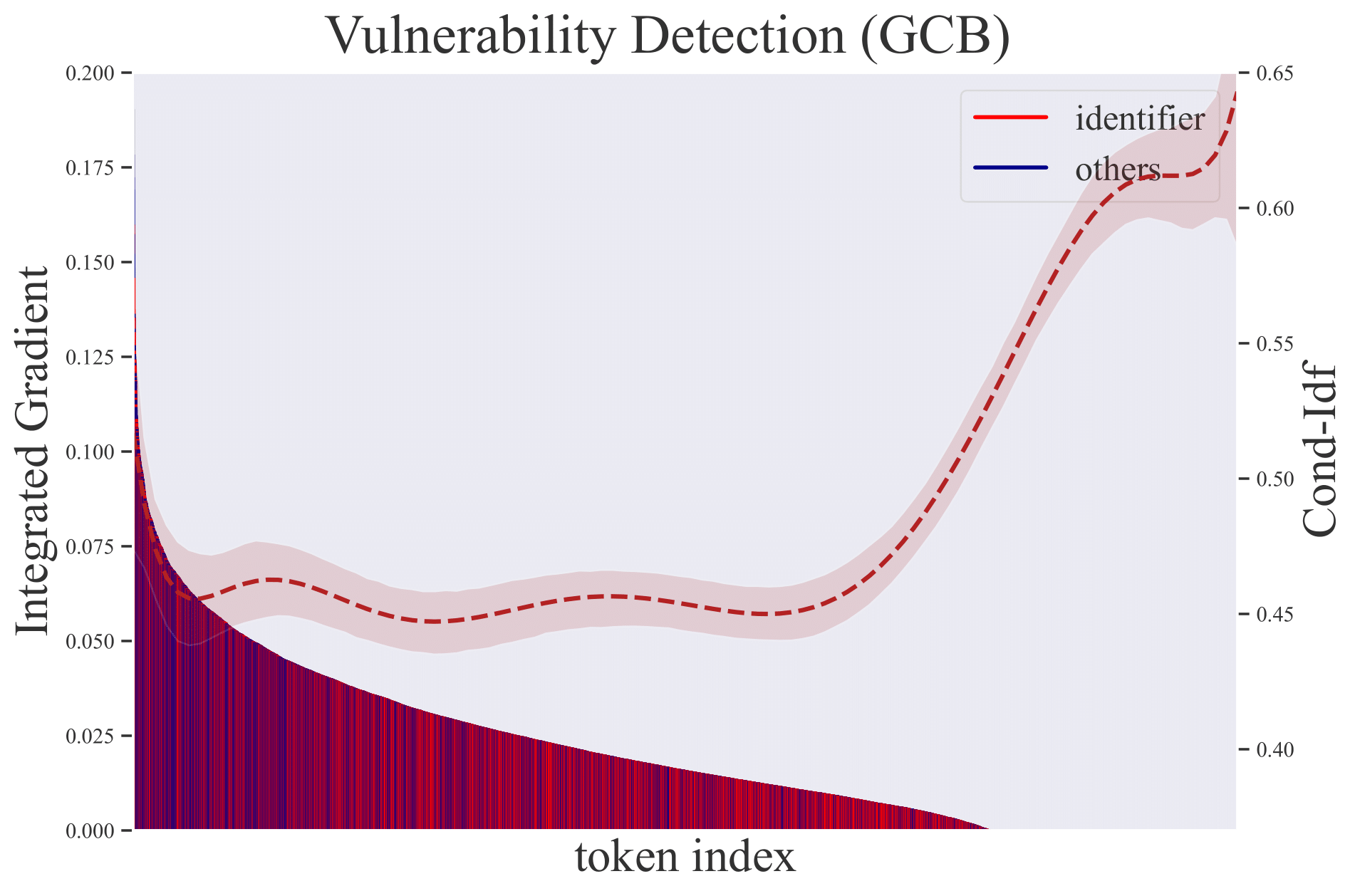}}
\subfigure{\label{fig:ti_ig_cb}\includegraphics[width=0.245\textwidth]{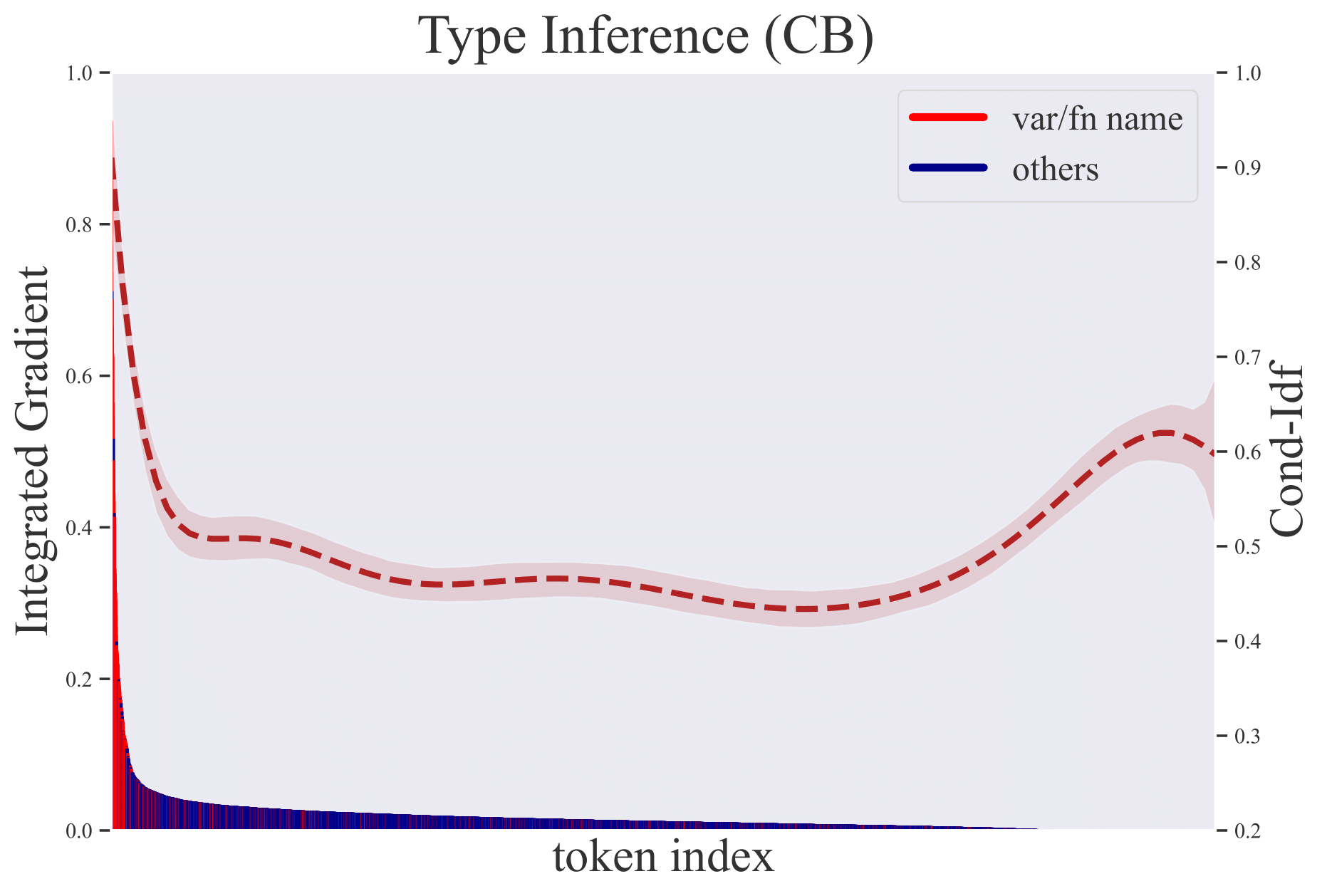}}
\subfigure{\label{fig:ti_ig_gcb}\includegraphics[width=0.245\textwidth]{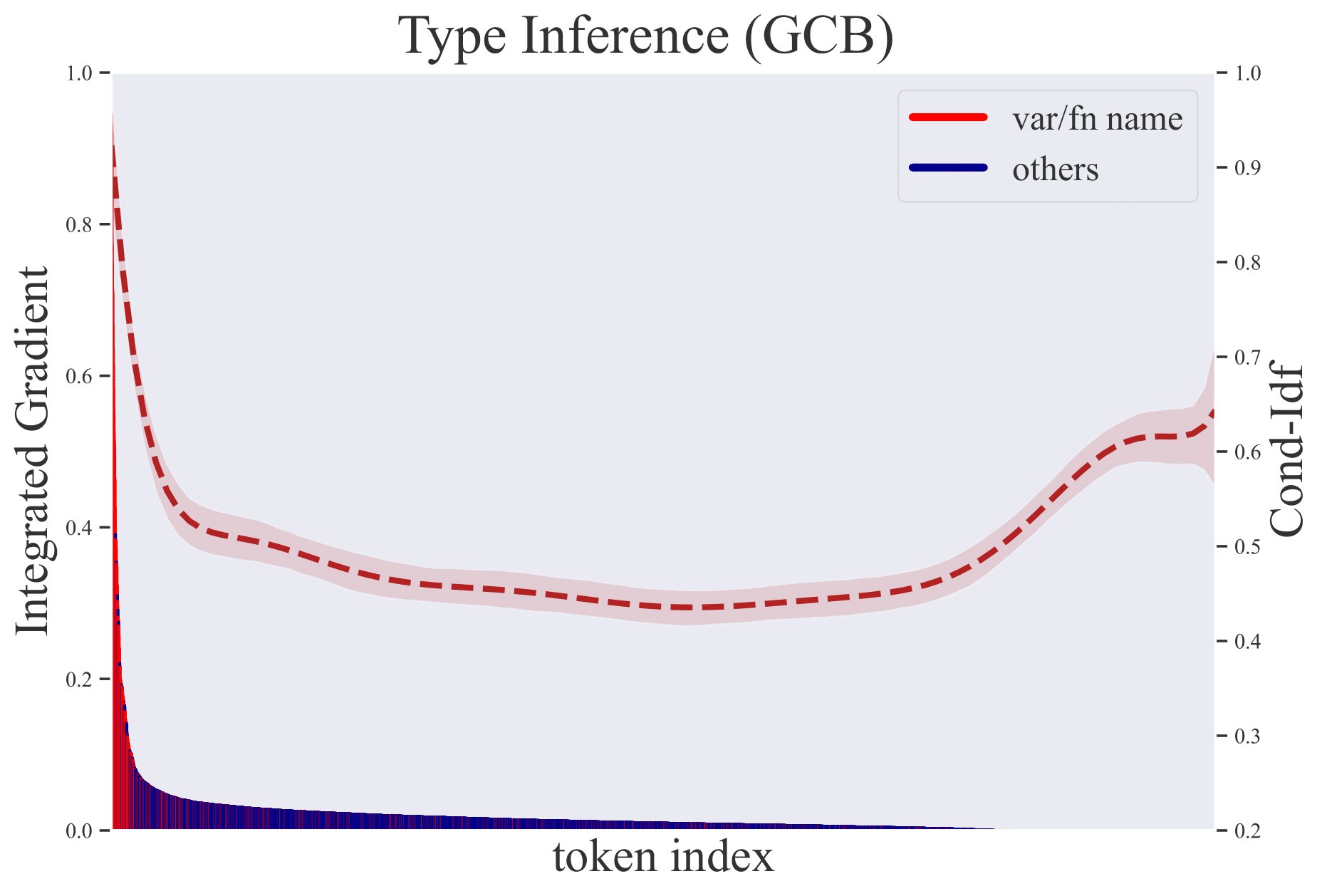}}\\
\subfigure{\label{fig:vd_abl_cb}\includegraphics[width=0.245\textwidth]{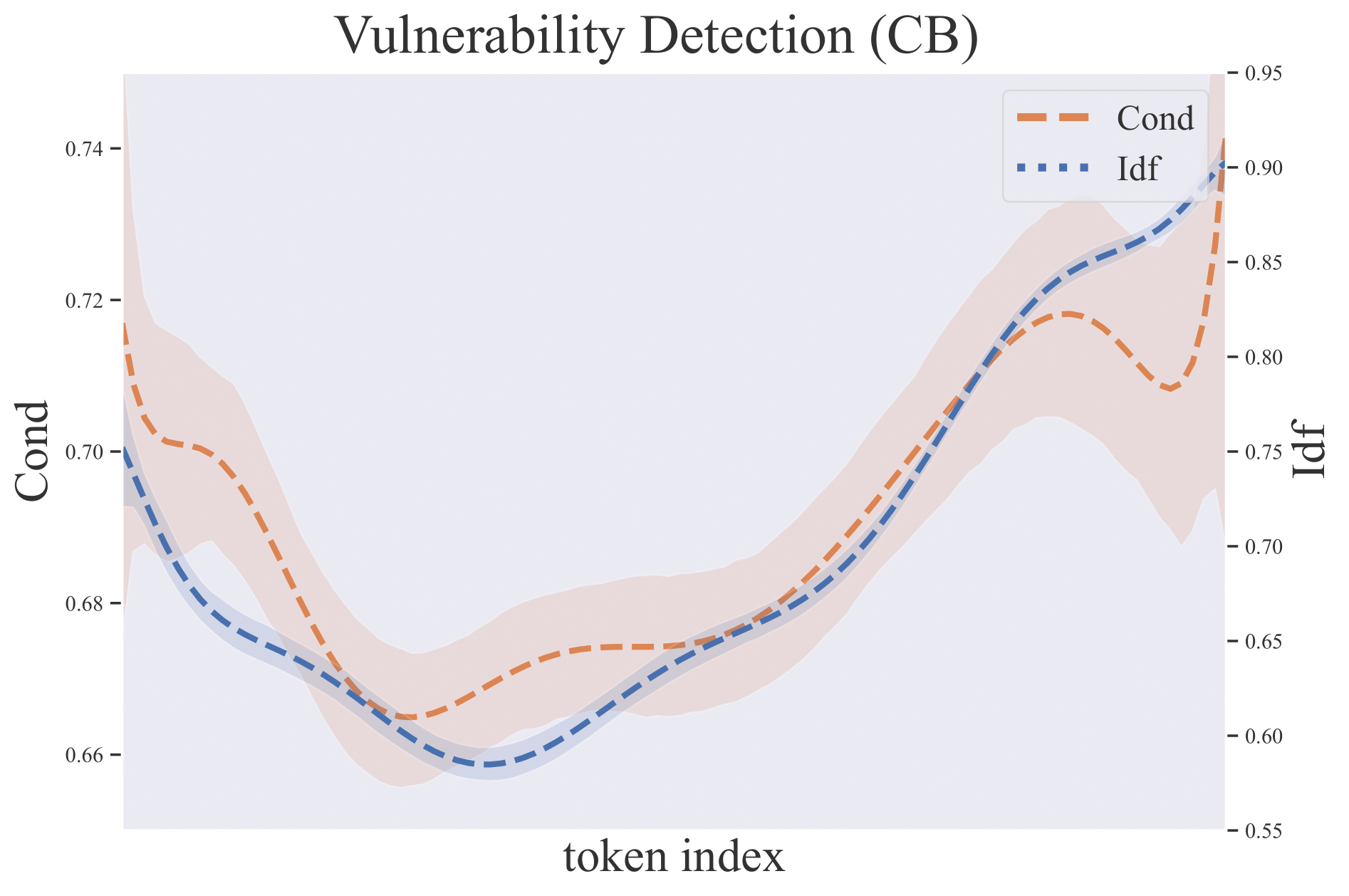}}
\subfigure{\label{fig:vd_abl_gcb}\includegraphics[width=0.245\textwidth]{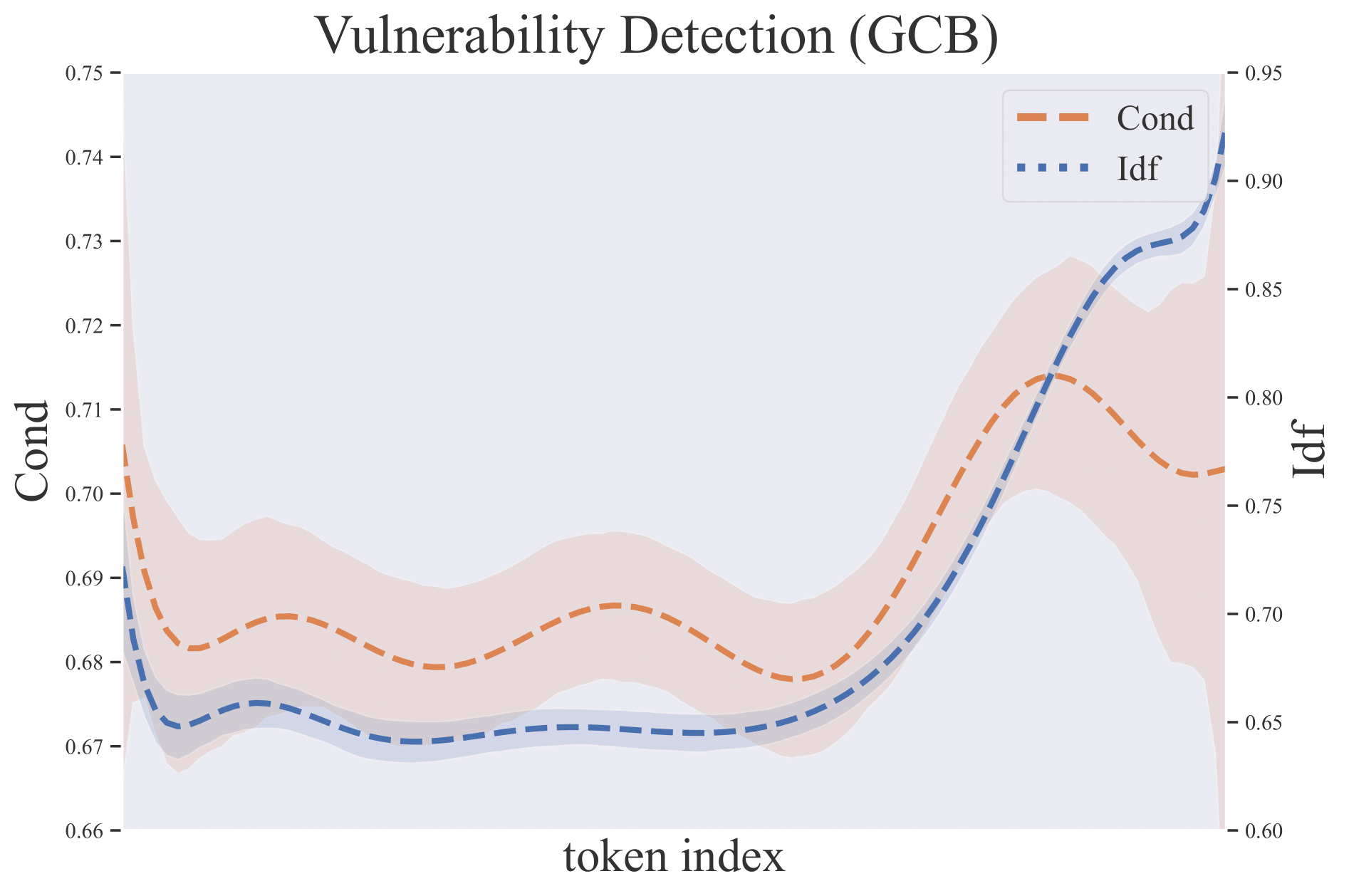}}
\subfigure{\label{fig:ti_abl_cb}\includegraphics[width=0.245\textwidth]{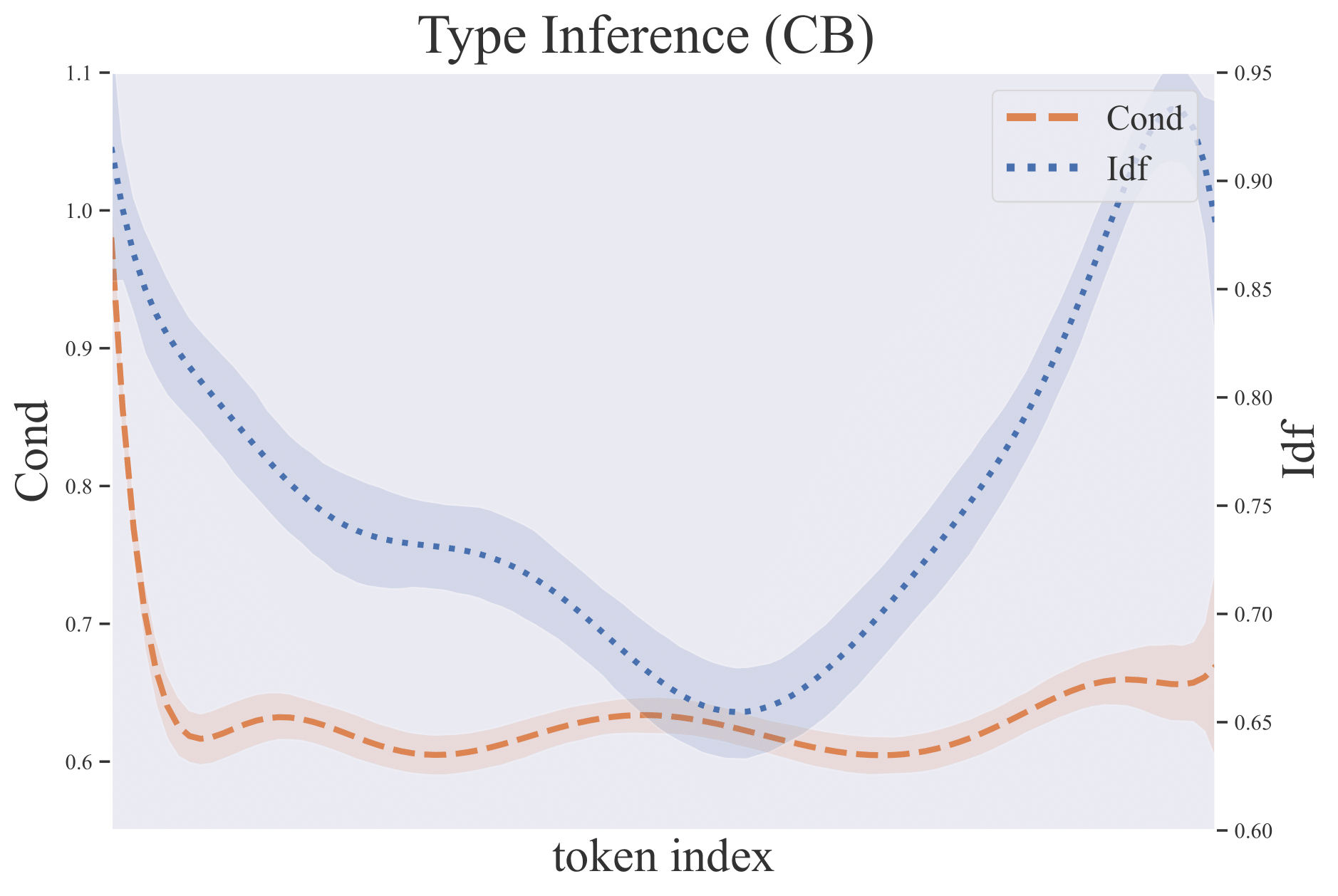}}
\subfigure{\label{fig:ti_abl_gcb}\includegraphics[width=0.245\textwidth]{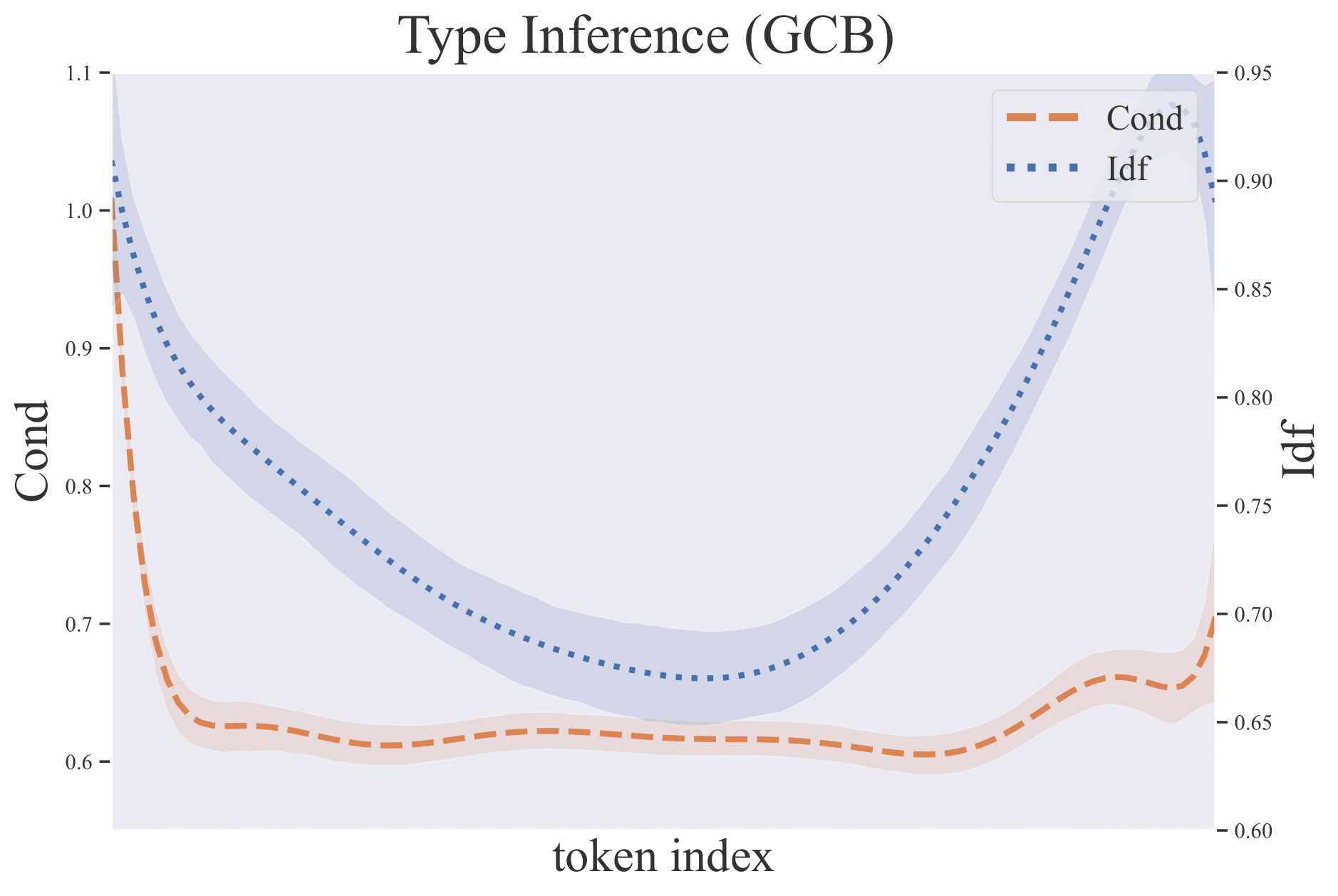}}

 \caption{For the figures in the first row: the lower distribution in each figure is the ranked integrated gradient distribution. The upper red dashed line denotes the polynomial regression approximation of the corresponding Cond-Idf distribution in terms of token index. CB and GCB denote the CodeBERT and GraphCodeBERT models. For the figures in the second row: The blue and orange lines denote the polynomial regression approximation of the Cond and Idf distribution in terms of token index respectively.}
\label{fig:ig_explain}
\vspace{-1em}
\end{figure*}
\paragraph{Dataset Unshuffling.} Given a training set $\mathcal{X}$, we first compute a similarity matrix $\mathbf{\lambda} \in \mathbb{R}^{ |\mathcal{X}| \times |\mathcal{X}|}$ of the training set (see line~1-3). Specifically, for samples $(x_{i}, x_{j})\in \mathcal{X} \times \mathcal{X}$, we first map them to a measure space with the embedding function $\mu$, then calculate their similarity with the similarity function $\kappa$: $\lambda_{ij}\leftarrow \kappa{(\mu(x_{i}), \mu(x_{j}))}$. Intuitively, this measures the level of logic closeness (similarity) of syntactic or semantic invariance between samples. For type inference, we focus on variables with assignment, we use bag-of-words (BoW)~\cite{harris1954distributional} vector that consists of tokens within the assignment statement of the target word as embedding $\mu$ and use cosine similarity $\mathrm{S_c}$ as $\kappa$ to calculate logic closeness: $\kappa(\mu(\cdot),\mu(\cdot))=\mathrm{S_c}(BoW(ASSIGN(\cdot)),BoW(ASSIGN(\cdot)))$, where $ASSIGN(\cdot)$ denotes the function that extracts the body of the assignment statement of the variable. The intuition is that the body of the assignment statement often contains consistent and generalizable information that is related to the type of the variable. For vulnerability detection, we use an encoder model pre-trained only on adversarial samples as $\mu$ (we use CodeBERT in this work). The adversarial samples are constructed from semantic-preserving identifier name replacement, denoted as $\rho$ (refer to Section \ref{task}). The intuition is that this encoder would only rely on the generalizable code semantics for detection as the biased user-defined token names are all normalized. The similarity score between a sample pair $(x_i, x_j)$ is calculated as: $\mathbf{\lambda}_{ij}=\mathrm{S_c}(\epsilon(\rho(x_i)), \epsilon(\rho(x_j)))$, where $\epsilon(\cdot)$ denotes the pre-trained embedding model, $\mathrm{S_c}$ denotes cosine similarity function. The training samples are then unshuffled and sorted according to the vectorized similarity matrix $sorted(vec(\mathbf{\lambda} \backslash \{\mathbf{\lambda}_{ij}|i=j\}))$ (see line~4-7). In this way, after batchifying, each mini-batch $B\in batches(\mathcal{X_{\mu}})$ would consist of samples with the closest logic relations.
\paragraph{In-batch Regularization.} Finally, during training, within each mini-batch, we use the cosine embedding loss $\mathbbm{1}^{ij}\mathrm{S_c}{(f(x_i; \theta), f(x_j; \theta))}$ to regularize the model into using similar representation when embedding samples with the same label and close logic relations. Here, $f$ denotes the backbone encoder, $\mathbbm{1}^{ij}$ is a boolean operator that selects samples with the same label. We weigh the loss of each sample pair with their corresponding similarity score $\lambda_{ij}$ to prevent misalignment. Detailed BPR loss is as follows:
\begin{equation}
\begin{aligned}
    \mathcal{L}_\mathrm{BPR}
    &=\mathbb{E}_{(x_i, x_j) \sim B\times B}{\mathbbm{1}^{ij}{\kappa(\mu(x_i), \mu(x_j))}}\\ 
    & \cdot \mathrm{S_c}{(f(x_i; \theta), f(x_j; \theta))}.
\end{aligned}
\end{equation}
$\mathcal{L}_\mathrm{BPR}$ can be trained together with existing mitigation methods and our final loss function is: $\mathcal{L}= \mathcal{L}_\mathrm{DEBIAS} + \gamma\cdot \mathcal{L}_\mathrm{BPR}$. In this work, we combine BPR with adversarial training/gradient reversal methods.
\begin{table}[t]
\scriptsize
\centering
\begin{tabular}{ccccccc} 
\toprule
               & \multicolumn{3}{c}{\textbf{CodeBERT}} & \multicolumn{3}{c}{\textbf{GraphCodeBERT}}  \\ 
\cmidrule(lr){2-4}\cmidrule(r){5-7}
\#Words        & Top1   & Top2   & Top3                   & Top1   & Top2   & Top3                   \\ 
\midrule
\textbf{Ratio}  & 66.6\% & 83.0\% & 90.0\%  & 13.2\% & 21.4\% & 28.8\%               \\
\bottomrule
\end{tabular}
\caption{ The ratio of samples whose top-$n$ ($n\in\{1, 2, 3\}$) interpretation words contain user-defined identifiers (vulnerability detection).}
\label{sample_level_vd}
\vspace{-1em}
\end{table}
$\mathcal{L}_\mathrm{DEBIAS}$ denotes loss function for adversarial training~\cite{yefet2020adversarial}/gradient reversal~\cite{stacey2020avoiding} (see Section~\ref{baseline}). The hyperparameter $\gamma$ denotes the regulatory coefficient for BPR loss. 


\section{Experiments}

\subsection{Experimental Setup}
\label{task}
\paragraph{Tasks \& Datasets.} We experiment with two representative program analysis tasks: 

\begin{itemize}[leftmargin=*]
  \item \emph{Type Inference}: 
  The goal of this task is to predict the type for variables/parameters/functions in a code snippet written in optionally typed language. In this work, we use the TypeScript dataset~\cite{hellendoorn2018deep}. 
  After preprocessing, the dataset contains samples from 233 TypeScript projects. We follow previous work~\cite{hellendoorn2018deep} and split the dataset by project into 80\%, 10\% and 10\% for inter-project OOD/adversarial training validation and test set. The training portion is also randomly shuffled and split by 80-20 proportions for IID train and test. We perform a semantic-preserving non-targeted attack~\cite{yefet2020adversarial} by replacing variable/parameters/function names with a set of dummy variables \eg~\{\texttt{var\_0}, \texttt{var\_1}, \texttt{...}\} on the corresponding inter-project samples to form the adversarial set.
  \item \emph{Vulnerability Detection}: 
  Previous work formulates vulnerability detection as a sequence/graph classification task, in which given a code snippet, the neural model should learn to predict whether it contains vulnerability or not. 
  In this work, we collect 999 C-language open-source projects from GitHub that contain vulnerabilities via keyword filtering in the commit message. We split the dataset by project into 70\% for IID training and testing 
  (randomly split by 80-20 proportions), 10\% and 20\% for inter-project OOD/adversarial validation and test set.We use the same adversarial set construction as the type inference task. 
\end{itemize}

\begin{table}[t]
\scriptsize
\centering
\begin{tabular}{ccccccc} 
\toprule
               & \multicolumn{3}{c}{\textbf{CodeBERT}} & \multicolumn{3}{c}{\textbf{GraphCodeBERT}}  \\ 
\cmidrule(lr){2-4}\cmidrule(r){5-7}
\#Words        & Top1   & Top2   & Top3                   & Top1   & Top2   & Top3                   \\ 
\midrule
\textbf{Ratio}  & 25.6\% & 31.6\% & 34.9\%  & 27.5\% & 33.0\% & 36.9\%               \\
\bottomrule
\end{tabular}
\caption{The ratio of samples whose top-$n$ ($n\in\{1, 2, 3\}$) interpretation words contain user-defined identifiers (type inference). }
\label{sample_level_ti}
\vspace{-1em}
\end{table}

\begin{table*}[t]
\centering
\scalebox{0.7}{
\begin{tabular}{ccccccc|cccccc} 
\toprule
\multicolumn{1}{l}{}    & \multicolumn{6}{c|}{\textbf{CodeBERT}}                                                                                         & \multicolumn{6}{c}{\textbf{GraphCodeBERT}}                                                                                                                                           \\ 
\cmidrule(l){2-13}
\multirow{2}{*}{\textbf{Methods}} & \multicolumn{3}{c}{\textbf{\textbf{Top-1 Acc (\%)}}} & \multicolumn{3}{c|}{\textbf{\textbf{\textbf{\textbf{Top-5 Acc (\%)}}}}} & \multicolumn{3}{c}{\textbf{\textbf{\textbf{\textbf{Top-1 Acc (\%)}}}}} & \multicolumn{3}{c}{\textbf{\textbf{\textbf{\textbf{\textbf{\textbf{\textbf{\textbf{Top-5 Acc (\%)}}}}}}}}}  \\
                        & INTRA & INTER & ADV                                  & INTRA & INTER & ADV                                                     & INTRA & INTER & ADV                                                    & INTRA & INTER & ADV                                                                                         \\ 
\cmidrule(lr){1-13}
Original                & 94.82 & 82.91 & 66.01                                & 98.35 & 94.87 & 86.00                                                   & 94.54 & 83.02 & 65.32                                                  & 98.77 & 95.05 & 84.53                                                                                       \\ 
\cmidrule(lr){1-13}
Reweighting             & 94.19 & 83.22 & 66.81                                & 98.73 & 95.65 & 86.24                                                   & 94.74 & 83.63 & 66.24                                                  & 98.97 & 95.88 & 84.76                                                                                       \\
PoE                     & 91.66 & 80.24 & 64.50                                 & 96.74 & 89.74 & 82.21                                                   & 93.84 & 82.56 & 64.73                                                  & 97.62 & 90.29 & 82.48                                                                                       \\ 
\cmidrule(lr){1-13}
Grad-rev                & 94.34 & 83.16 & 66.26                                & 98.32 & 95.28 & 85.73                                                   & 94.62 & 83.58 & 66.01                                                  & 98.69 & 95.63 & 84.38                                                                                       \\
Grad-rev w/ BPR         & 94.90  & 83.95 & 66.51                                & 98.85 & 95.84 & 86.77                                                   & 95.37 & 84.42 & 66.20                                                   & 98.93 & 96.06 & 84.82                                                                                       \\ 
\cmidrule(lr){1-13}
Adv-train               & 94.92 & 83.42 & 79.55                                & 98.84 & 96.08 & 94.27                                                   & 95.45 & 83.74 & 79.32                                                  & 99.08 & 95.82 & 94.06                                                                                       \\
Adv-train w/ BPR        & \colorbox{lightgray}{95.25} & \colorbox{lightgray}{84.45} & \colorbox{lightgray}{80.57}                                & \colorbox{lightgray}{98.87} & \colorbox{lightgray}{96.24} & \colorbox{lightgray}{94.31}                                                   & \colorbox{lightgray}{95.68} & \colorbox{lightgray}{84.95} & \colorbox{lightgray}{80.44}                                                  & \colorbox{lightgray}{99.12} & \colorbox{lightgray}{96.08} & \colorbox{lightgray}{95.06}                                                                                       \\
\bottomrule
\end{tabular}
}
\caption{Generalization and robustness evaluation on the type inference task.}
\label{ti_result}
\vspace{-1em}
\end{table*}

\paragraph{Mitigation Baselines.}\label{baseline} We evaluate BPR along with four representative mitigation baselines: two model-agnostic methods (reweighting, product-of-expert (PoE)) and two representation-based methods (adversarial training, gradient reversal). For reweighting~\cite{schuster2019towards,clark2019don}, it first obtains a biased model by training the model only on the biased features, then the output probability of the bias-only model $p_b$ is used to adjust the weights of the training samples to train the debiased model, such that the contribution of samples to which the biased model assigns high prediction probability is lower weighted.
For the PoE~\cite{he2019unlearn,clark2019don,mahabadi2020end}, it also requires a trained biased model, the debiased model is trained by ensembling its output probability with that of the biased model. 
For adversarial training~\cite{madry2017towards,yefet2020adversarial}, it optimizes the model based on both original samples and perturbed adversarial samples so that the co-occurrence of spurious data cues and labels are down-weighted. Finally, for gradient reversal~\cite{stacey2020avoiding,kim2019learning,minervini2018adversarially}, it unlearns the bias in a minimax game by predicting the target bias using model representation and reverses its gradient during backpropagation. 



\paragraph{Implementation Details.} 
We focus on the analysis of two encoder-only LLMs-based neural code models: the pretrained CodeBERT (CB) and GraphCodeBERT (GCB) models. We use the open-sourced checkpoints from Feng~\etal~\cite{feng2020codebert} and Guo~\etal~\cite{DBLP:conf/iclr/GuoRLFT0ZDSFTDC21}. For both benchmarks, we fine-tune the model for 10 epochs, which all models could converge. We use the Adam optimizer for the update and the learning rate is set as $2*10^{-5}$. The training batch size is set as 16. 
We conducted all experiments on a Ubuntu 18.04 server with 24 cores of 2.20GHz CPU, 251GB RAM
and two Quadro RTX 8000 GPUs.

\subsection{Project-Specific Bias Analysis \& Interpretation}
In this section, we quantitatively analyze and interpret the project-specific bias learning behavior of the CB and GCB models.
\begin{table*}[t]
\centering
\scalebox{0.7}{
\begin{tabular}{ccccccc|cccccc} 
\toprule
\multicolumn{1}{l}{}              & \multicolumn{6}{c|}{\textbf{CodeBERT}}                                                                                        & \multicolumn{6}{c}{\textbf{GraphCodeBERT}}                                                                                                                                          \\ 
\cmidrule(l){2-13}
\multirow{2}{*}{\textbf{Methods}} & \multicolumn{3}{c}{\textbf{\textbf{Top-1 Acc (\%)}}} & \multicolumn{3}{c|}{\textbf{\textbf{\textbf{\textbf{F1-score (\%)}}}}} & \multicolumn{3}{c}{\textbf{\textbf{\textbf{\textbf{Top-1 Acc (\%)}}}}} & \multicolumn{3}{c}{\textbf{\textbf{\textbf{\textbf{\textbf{\textbf{\textbf{\textbf{F1-score (\%)}}}}}}}}}  \\
                                  & INTRA & INTER & ADV                                  & INTRA & INTER & ADV                                                    & INTRA & INTER & ADV                                                    & INTRA & INTER & ADV                                                                                        \\ 
\cmidrule(lr){1-13}
Original                          & 81.61 & 64.01 & 61.50                                 & 82.37 & 67.94 & 67.83                                                  & 81.70  & 64.48 & 64.34                                                  & 83.53 & 68.51 & 68.18                                                                                      \\ 
\cmidrule(lr){1-13}
Reweighting                       & 80.34 & 61.99 & 58.64                                & 81.06 & 67.35 & 67.31                                                  & 80.67 & 63.78 & 59.93                                                  & 81.85 & 67.49 & 66.2                                                                                       \\
PoE                               & 81.24 & 63.09 & 58.12                                & 81.90  & 68.02 & 67.73                                                  & 81.28 & 63.61 & 61.37                                                  & 83.04 & 68.39 & 67.25                                                                                      \\ 
\cmidrule(lr){1-13}
Grad-rev                          & 81.66 & 64.52 & 63.21                                & 83.05 & 69.00    & 68.82                                                  & 81.27 & 64.17 & 64.07                                                  & 82.78 & 69.86 & 69.78                                                                                      \\
Grad-rev w/ BPR                   & 81.73 & 64.57 & 64.56                                & 83.20  & 69.21 & 69.38                                                  & 81.57 & 65.12 & 65.01                                                  & 82.33 & 70.05 & 70.02                                                                                      \\ 
\cmidrule(lr){1-13}
Adv-train                         & 81.89 & 64.53 & 63.31                                & 82.60  & 68.92 & 68.80                                                   & 81.91 & 64.95 & 64.98                                                  & 84.07 & 68.89 & 68.89                                                                                      \\
Adv-train w/ BPR                  &\colorbox{lightgray}{82.76} & \colorbox{lightgray}{64.72} & \colorbox{lightgray}{64.83}                                & \colorbox{lightgray}{83.24} & \colorbox{lightgray}{69.37} & \colorbox{lightgray}{69.41}                                                  & \colorbox{lightgray}{82.46} & \colorbox{lightgray}{66.31} & \colorbox{lightgray}{66.38}                                                  & \colorbox{lightgray}{84.47} & \colorbox{lightgray}{70.18} & \colorbox{lightgray}{70.18}                                                                                      \\
\bottomrule
\end{tabular}
}
\caption{Generalization and robustness evaluation on the vulnerability detection task.}
\label{vd_result}
\end{table*}
\paragraph{Bias Behavior Analysis.}
We calculate the mean integrated gradient for each token in the IID test set vocabulary and rank them in descending order to obtain the sorted distribution. Then, for vulnerability detection, we perform lexical analysis and categorize the vocabulary into user-defined identifiers (denoted as identifier) and others. The identifier category contains user-defined variable/function names and macro-definition names. As shown in Figure~\ref{fig:ig_explain} (first row), we illustrate the integrated gradient weights of the project-specific tokens with red bars and others with blue bars. 
For type inference, we categorize tokens into user-defined Boolean variable/function name identifiers and others. Similarly, we use red and blue bars to represent the integrated gradient weights of these two categories, as shown in Figure~\ref{fig:ig_explain} (first row).
As shown in the distribution, we observe that the area under the head of the distribution is heavily reddish for both models in terms of the type inference task, which indicates that the models focus heavily on user-defined components. Quantitatively, we take the top 1\% of the tokens as the head. While the user-defined identifiers only occupy 15.0\% of the vocabulary, 88.1\% of tokens within the head fall into this category for the CB model, and 83.6\% for the GCB model. The results indicate that the data-flow aware pretraining objectives of the GCB model allow it to be less biased than the CB model on the type inference task which is reliant on explicit def-use relations. 
The result is consistent for the vulnerability detection task. With the top 1\% of the tokens as the head, while the self-defined tokens take up 56.6\% of the vocabulary, 66.8\% of the tokens within the head belong to the user-defined components for the CB model, while it drops to 39.7\% for the GCB model.
Furthermore, we measure the extent of the bias learning behavior from the sample level. 
Specifically, we calculate the ratio of samples whose top-$n$ ($n\in\{1, 2, 3\}$) integrated gradient tokens contain user-defined identifiers. The results are shown in \Cref{sample_level_vd},\ref{sample_level_ti}. The results indicate that the behavior of both the CB and GCB models are biased for a considerable amount of samples. For example, for vulnerability detection (CB), up to 90.0\% of the samples leverage project-specific user-defined tokens as its top-3 attribution words for prediction. 

\paragraph{Bias Behavior Interpretation.}
Given the ranked integrated gradient distribution, we calculate the corresponding Cond-Idf distribution in terms of the token index and approximate it with polynomial regression (order=10). As shown in the first row of \Cref{fig:ig_explain}, we observe that for the head part of the distribution where models give relatively high attribution weights, the fitted curve of Cond-Idf is positively correlated with the integrated gradient distribution for both tasks. 
We further conduct a detailed ablation analysis of the Cond-Idf measurement, the results are shown in the second row of \Cref{fig:ig_explain}. Specifically, we break the measurement into the conditional probability distribution approximation (Cond) and Idf distribution approximation. 
For the two evaluated models on both the vulnerability detection and type inference tasks, we can observe that the head of the ranked integrated gradient distribution positively correlates with both the Cond and Idf distributions. Quantitatively, to calculate the correlation, we sample 50 data points that are evenly spaced from the top 50\% of both the ranked integrated gradient distribution and the Cond/Idf distributions. We use the Spearman’s rank correlation as the measurement. \Eg~for the GCB model on the type inference task, the head of the Cond measurement and the ranked integrated gradient distribution are correlated with a Spearman's rank correlation of $\rho_\mathrm{{Cond}}=0.566$; and for the Idf measurement: $\rho_\mathrm{{Idf}} =1.00$. Similarly, for the CB model on the vulnerability detection task, the two measurements correlate with the ranked integrated gradient distribution with high rank correlation ($\rho_\mathrm{{Cond}}=0.561$, $\rho_\mathrm{{Idf}} =0.700$).

\subsection{Bias Mitigation Effectiveness}
\begin{figure}[t]
\centering
\includegraphics[width=0.42\textwidth]{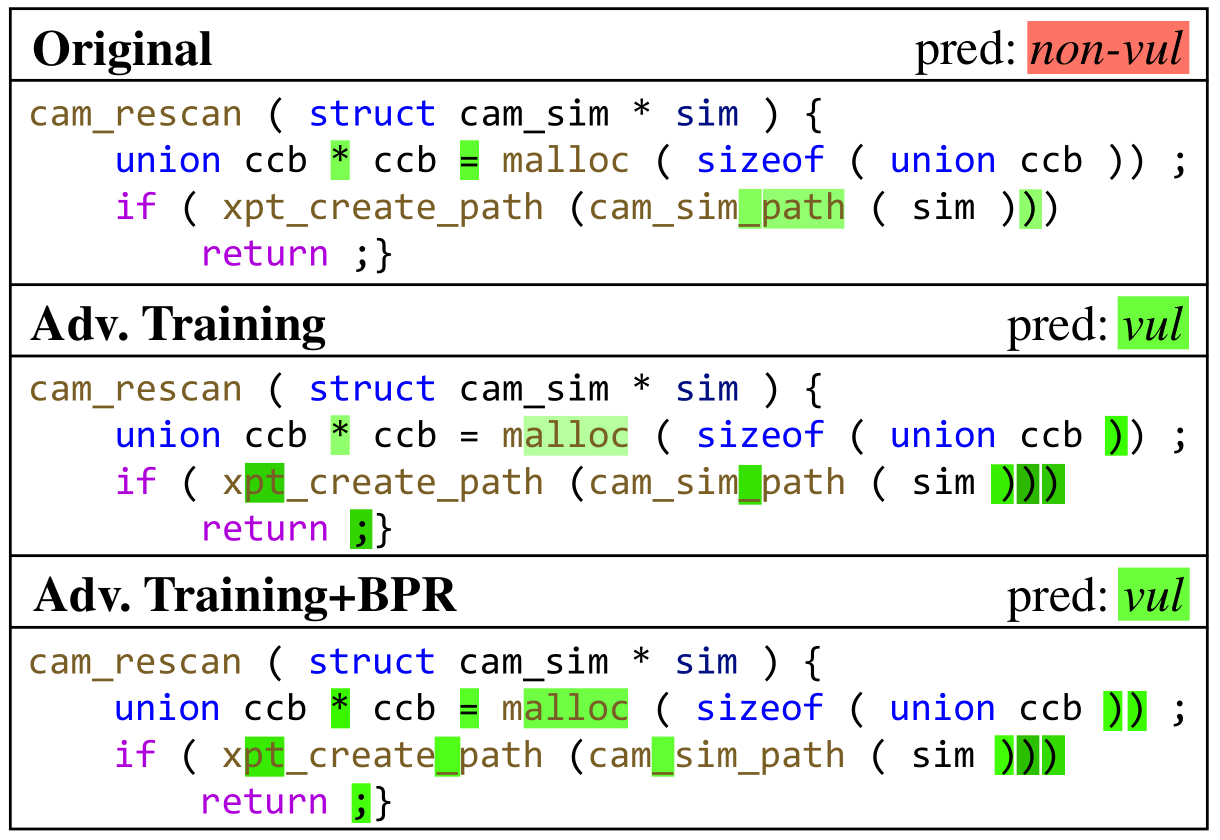}
\caption{Illustrated example of applying BPR compared to the baseline methods for the vulnerability detection task.}
\label{case_study2}
\end{figure}

We present results on the intra-project IID test set, inter-project OOD set and adversarial set. 
In \Cref{ti_result}, we evaluate the type inference performance in terms of top-1 and top-5 accuracy following previous works~\cite{wei2020lambdanet,jesse2021learning}. 
In Table~\ref{vd_result}, we evaluate the vulnerability detection performance in terms of top-1 accuracy and F1-score.
We have the following three key findings: 

\begin{itemize}[leftmargin=*]
    \item With the standard IID training-test split, both models achieve decent performance on the intra-project IID test set relying on shortcut features, whereas their performance drops significantly on the inter-project OOD and adversarial set. For example, for type inference, CB achieves 94.82\%@top1-Acc on the intra-project test set and drops to 82.91\% on the inter-project OOD and further decreases to 66.01\% on adversarial data. The results indicate that the bias learning behavior seriously undermines models' generalizability and robustness. 

    \item Among the baselines evaluated, model-agnostic mitigation approaches (reweighting and PoE) are less helpful compared to representation-based methods (adversarial training and gradient reversal). For example, for vulnerability detection (CB), adversarial training increases adversarial robustness accuracy by +1.81\%@top1-Acc, while reweighting and PoE decrease it by -2.86\% and -3.38\% respectively. The results are similar for type inference. 
    One possible reason is that there is no clear boundary for defining biased and unbiased sample, since every code sample requires the usage of user-defined words like variable/ function names, thus representation-based mitigation methods are more effective compared with model-agnostic methods. 
    \item When combined with Grad-rev and Adv-train, BPR demonstrates a consistent capacity to enhance both generalization and robustness by learning more robust representation. Notably, it is intriguing to observe that the implementation of BPR can even yield enhancements in terms of IID performance. For example, for vulnerability detection (GCB), BPR improves adversarial training by +1.40\%@top1-Acc, +1.29\%@F1 on the adversarial set. 
    To better understand the source of improvement, we compute the mean integrated gradient values of the ground-truth tokens that the model should focus on for the original model, model trained with adversarial training and model trained with adversarial training w/ BPR (we experiment on the GCB model since it achieves overall better performance). For vulnerability detection, we compute the mean IG value of the memory management API 
    \begin{figure}[t] 
\centering 
\includegraphics[width=0.44\textwidth]{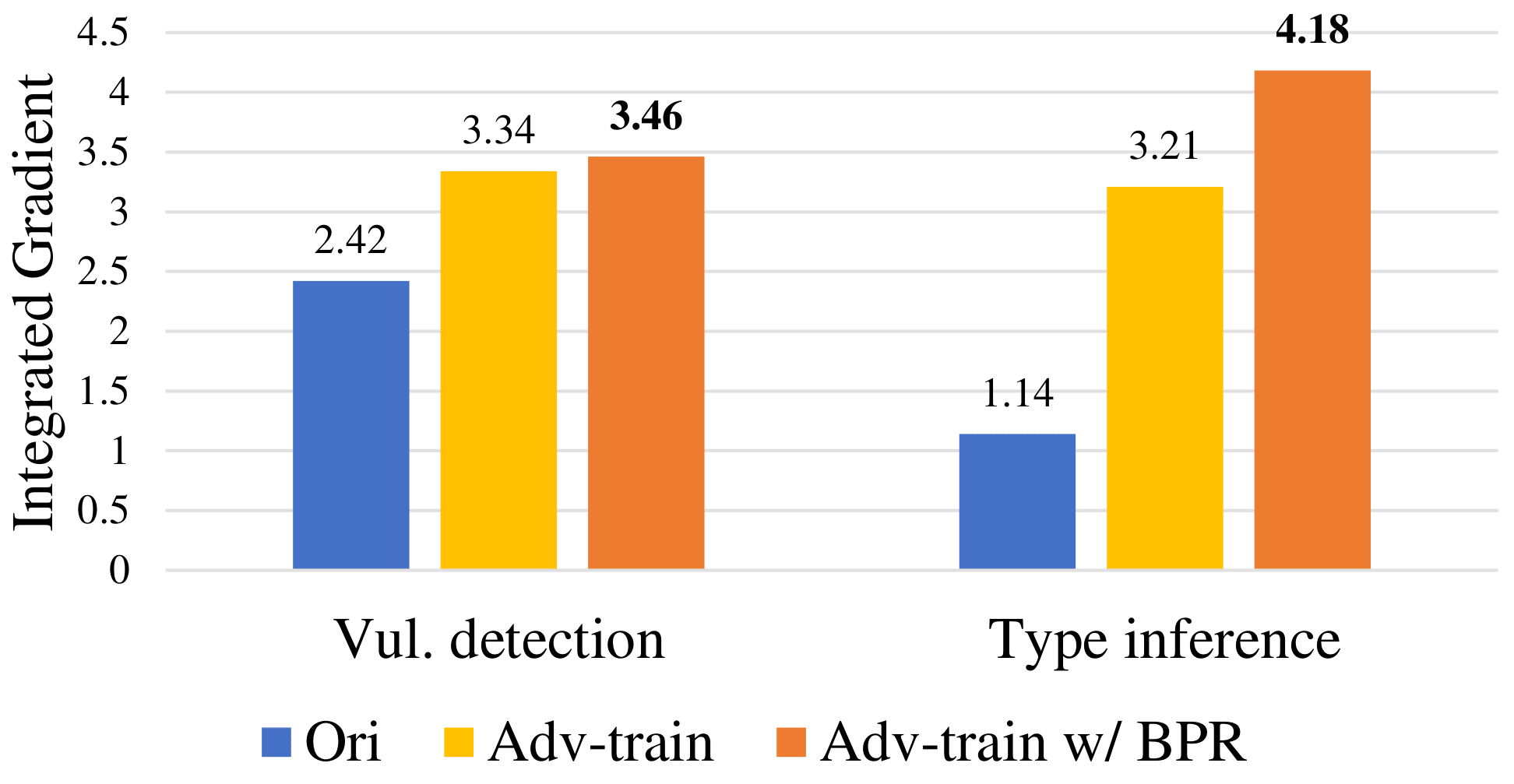} 
\caption{Mean integrated gradient of the original and mitigated GCB models on the ground-truth evidence.} 
\label{ig_ablation} 
\end{figure}
    of the C standard library. And for type inference, we compute the value of the boolean constant and logical operators. The results are shown in \Cref{ig_ablation}. Specifically, adversarial training significantly improves over the original GCB model, and by incorporating BPR, the mean IG values of the ground-truth tokens are all further increased, which indicates that the model's behavior is much more robust and focuses more on the ground-truth tokens. We further conduct a case study of vulnerability detection as shown in \Cref{case_study2}. The union \texttt{ccb} is first allocated with memory using the C standard API \texttt{malloc}. However, it would cause a memory leak as the program directly return while failing to free up the memory properly. It is obvious that the original model completely ignores the ground-truth API token \texttt{malloc} and erroneously predicts the function as non-vulnerable. When trained with adversarial training, model starts focusing on it and infers correctly. Finally, when incorporated with BPR, model robustly predicts it as vulnerable by paying much higher attention to the ground-truth evidence.
\end{itemize}

\section{Related Work}
\paragraph{Shortcut Learning \& Mitigation.} DNNs have been shown powerful and prevalent in many areas \cite{he2015deep,vaswani2017attention,hu2021enumeration,Hu_2023,huang2023zero}. However, researchers observe that neural models tend to leverage shallow statistical cues instead of generalizable features for prediction \cite{du2023shortcut, li2023fairer,tianlin2023runner}. For the VQA task, it is found that the model often conditions its predictions on language prior while ignoring the image~\cite{manjunatha2019explicit,agrawal2018don}. 
For the NLI task, models tend to focus on a single branch of input or frequent but spurious unigram words~\cite{schuster2019towards,niven2019probing}. One of the most representative line of mitigation method is debiasing from bias-only model, which includes re-weighting~\cite{schuster2019towards}, product-of-expert~\cite{he2019unlearn,clark2019don,zhou2020towards,cadene2019rubi}, 
etc. Representation-based methods are another line of effort that is proven to be effective. The idea is to orthogonalize model's representation from the bias features~\cite{madry2017towards,stacey2020avoiding,kim2019learning}.


\paragraph{Code Representation Learning.} Neural models are useful in modeling programming language data and perform well in software analysis tasks, \eg vulnerability detection~\cite{zhou2019devign}, program synthesis~\cite{chen2018execution}, etc. 

The Transformer architecture-based Large Language Models (LLMs) have become the most widely used neural code models due to their state-of-the-art performance on a wide range of downstream tasks. 
These models can be categorized into three major groups according to their architectures~\cite{hou2023large}: encoder-only~\cite{feng2020codebert,DBLP:conf/iclr/GuoRLFT0ZDSFTDC21}, encoder-decoder~\cite{ahmad2021unified,raffel2020exploring,wang2021codet5}, and decoder-only~\cite{chen2021evaluating,touvron2023llama} models.
Many of the previous neural code models literature conduct evaluation only under the \emph{intra-project} setting instead of the \emph{inter-project} setting, despite the fact that the latter is closer to reality. In addition, previous work notices that these neural code models are vulnerable to naive adversarial attack~\cite{yefet2020adversarial} such as variable name change or dead code insertion. 
\section{Conclusion and Future Work}
In this work, we analyze the project-specific bias learning behavior of the encoder-only LLMs-based neural code models, which renders them ungeneralizable to inter-project OOD or adversarial settings. We observe that this phenomenon can be interpreted via the Cond-Idf measurement. Furthermore, we propose a general mitigation mechanism BPR that forces the model to infer based on robust representation using logic relations among samples. Experimental results on two representative benchmarks validate that BPR improves OOD generalization and adversarial robustness while not sacrificing IID performance. In the future, we plan to study bias learning behavior on more benchmarks and model architectures.

\section{Acknowledgments}
This research is supported by the National Research Foundation, Singapore, and the Cyber Security Agency under its National Cybersecurity R\&D Programme (NCRP25-P04-TAICeN) and NRF Investigatorship NRF-NRFI06-2020-0001. Any opinions, findings and conclusions or recommendations expressed in this material are those of the author(s) and do not reflect the views of National Research Foundation, Singapore and Cyber Security Agency of Singapore.
\section*{Limitations}
Despite that BPR can effectively increase the generalization and robustness of the model, one limitation of our approach is that it requires careful design of the modeling of human expert knowledge. However, for many software analysis tasks, human expert knowledge is suboptimal or difficult to abstract. Thus automatic identification of latent bias for neural code model would be an important future direction.
\section*{Ethics Statement}

We state that the vulnerability detection dataset we used in this work is collected from GitHub via keyword mapping in the commit message and has been through rigorous human review, in which all vulnerabilities are fully disclosed and repaired by developers, and shall contain no sensitive information or exposure of privacy. Thus, it would not produce any potential negative societal consequences.
\nocite{*}
\section{Bibliographical References}\label{sec:reference}

\bibliographystyle{lrec-coling2024-natbib}
\bibliography{lrec-coling2024-example}


\end{document}